# Penalty logic and its link with Dempster-Shafer theory

**Florence Dupin de Saint-Cyr,   Jérôme Lang**
IRIT - Université Paul Sabatier
118 Route de Narbonne
31062 Toulouse Cedex
France
{ dupin, lang } @irit.fr

**Thomas Schiex**
CERT-ONERA
2 Avenue Edouard Belin - BP 4025
31055 Toulouse Cedex
France
schiex@cert.fr

## Abstract

Penalty logic, introduced by Pinkas [17], associates to each formula of a knowledge base the price to pay if this formula is violated. Penalties may be used as a criterion for selecting preferred consistent subsets in an inconsistent knowledge base, thus inducing a non-monotonic inference relation. A precise formalization and the main properties of penalty logic and of its associated non-monotonic inference relation are given in the first part. We also show that penalty logic and Dempster-Shafer theory are related, especially in the infinitesimal case.

## 1 Introduction

The problem of *inconsistency handling* appears when the available knowledge base – KB for short – (here a set of propositional formulas) is inconsistent. Most approaches come up with the inconsistency by selecting among the consistent subsets of KB some *preferred* subsets; the selection criterion generally makes use of uncertainty considerations, sometimes by using explicitly uncertainty measures (such as Wilson [27], Benferhat and Smets [2]), or more often using measures expressed qualitatively as *priorities* (the idea comes back to Rescher [20] and has been developed by many authors, among them Brewka [3], Nebel [16], Cayrol [4], Benferhat, Cayrol, Dubois, Lang, Prade [1] and Lehmann [14]). Although these priorities are generally not given a semantics in terms of uncertainty measures (however see [1] for a comparative study of the priority-based and possibilistic approaches to inconsistency handling), their intuitive interpretation is clearly in terms of gradual uncertainty: the least prioritary formulas (*i.e.*, the ones which are most likely to be rejected in case of inconsistency) are clearly the ones we are the least confident in, *i.e.*, the least certain ones. All aforementioned priority-based approaches consist in ranking the $KB$ in $n$ priority levels (assume that 1 is the highest priority and $n$ the lowest) and maximize the set *or* the number of formulas satisfied at each level, with the condition that violating however many formulas at a given level is always more acceptable than violating only one formula at a strictly higher level: thus these approaches are *non-compensatory*, *i.e.*, levels never interact.

An alternative approach, more or less empirical but apparently very appealing (besides it has already been used several times in the literature) consists in weighting formulas with positive numbers called *penalties*. Contrarily to priorities, penalties are compensatory since they are *additive*: the global penalty for rejecting a set of formulas is the sum of the elementary penalties of the rejected formulas. Moreover, inviolable (or unrejectable) formulas are given an infinite penalty. The additive combination of penalties leads to an interpretation in terms of *cost*, thus this criterion is *utilitarist*, contrarily to priority-based approaches which are rather *egalitarist*. This additive criterion is very intuitive, since rejecting a formula generally causes some "additive" trouble with the experts which provided the $KB$ with the formulas, or some real financial cost, or another kind of additive cost. Note that a degenerate case of penalties (all penalties being equal to 1) prefers subsets of maximum cardinality. Moreover, and as we will see later, these penalties can sometimes be interpreted as the "probability of fault" of the source which provided us with the information (all sources failing independently), up to a logarithmic transformation. In any case, these penalties can be viewed as *measuring uncertainty* since, again, the less expensive to reject, the more uncertain the piece of information. Thus, penalty logic *expresses uncertainty in terms of costs*. However a formal connection of penalties with classical theories of uncertainty has not really been made.

Penalty-based approaches have been already used several times in the literature, first by Pinkas 91 [17] (from whom we borrowed the terminology "penalty") who uses them for inconsistency handling and for modelling symmetric neural networks behavior, and also by Eiter and Gotlob 94 [10] for cost-based abduction, by Sandewall 92 [21] for cost-based minimization of surprises in temporal reasoning and by Freuder and Wallace [12] for tackling inconsistencies in Constraint Satisfaction Problems. Moreover, penalties associated to formulas have also been used for guiding the search in randomized algorithms dedicated to the satisfiability problems, such as GSAT [23, 22]. Lastly, there should clearly be a link between penalties and utility theory (the latter has been recently used in AI, especially in decision-theoretic planning – see e.g. [18]); however, in this paper we do not investigate this possible link.

In this paper we revisit penalties by giving a further formalization of Pinkas' work; we also go further in the theoretical study of penalty-based inconsistency handling and non-monotonic reasoning. We briefly give a formalization in penalty logic of an additive O.R. problem. Lastly, we establish a link between penalties and Dempster-Shafer theory; this link is twofold: first, the penalty function (on interpretations) is equivalent, up to a log-transformation, to a contour function (i.e., the plausibility function restricted to singletons); then penalty functions on formulas coincide with plausibility functions of an *infinitesimal version* of Dempster-Shafer theory.

## 2 Penalty logic

### 2.1 Formal definitions

In the following, $\mathcal{L}$ will be a propositional language based on a finite number of propositional variables. $\top$ and $\bot$ will represent tautology and contradiction respectively. Formulas of $\mathcal{L}$ will be written $\varphi$, $\psi$, etc. The set of interpretations attached to $\mathcal{L}$ will be denoted by $\Omega$, and an interpretation by $\omega$. $\varphi \models \psi$ and $\varphi \dashv\vdash \psi$ will represent logical consequence and logical equivalence between the formulas $\varphi$ and $\psi$ respectively. $\models$ will also be used between an interpretation and a formula to denote satisfiability. The set of models of a formula $\varphi$ will be denoted by $M(\varphi)$; the set of formulas of $\mathcal{L}$ satisfied by $\omega$, i.e., $\{\varphi \mid \omega \models \varphi\}$ will be denoted by $[\omega]$.

A classical knowledge base $\mathcal{B}$ is a set of formulas of $\mathcal{L}$. A sub-theory of $\mathcal{B}$ is a consistent subset of $\mathcal{B}$. A maximal sub-theory $T$ of $\mathcal{B}$ is a consistent subset of $\mathcal{B}$ such that $\forall \varphi \in \mathcal{B} \setminus T, T \cup \{\varphi\}$ is inconsistent. Given a formula $\psi$, $T$ is said $\psi$-consistent iff $T \cup \{\psi\}$ is consistent; $T$ is maximal $\psi$-consistent if it is $\psi$-consistent and $\forall \varphi \in \mathcal{B} \setminus T, T \cup \{\varphi, \psi\}$ is inconsistent. $\overline{\mathbb{R}}^{\star +}$ will be the union of the set of all the strictly positive real numbers and $\{+\infty\}$, equipped with the usual order (in particular, if $\alpha \neq +\infty$ then $\alpha < +\infty$).

A *penalty knowledge base* PK is a finite multi-set of pairs $\langle \varphi_i, \alpha_i \rangle$ where $\varphi_i \in \mathcal{L}$ and $\alpha_i \in \overline{\mathbb{R}}^{\star +}$. $\alpha_i$ is the penalty associated to $\varphi_i$; it represents intuitively what *we should pay in order to get rid of* $\varphi_i$, if we pay the requested price we do not need any longer to satisfy $\varphi_i$; so the larger $\alpha_i$ is, the more important $\varphi_i$ is.

In particular, if $\alpha_i = +\infty$ then it is forbidden to remove $\varphi_i$ from PK ($\varphi_i$ is *inviolable*).

Since PK is a multi-set of pairs (and not a set), it is possible for a pair $\langle \varphi, \alpha \rangle$ to appear several times in PK; for example, PK $= \{\langle a, 1\rangle, \langle a, 1\rangle\}$ is not equivalent to PK' $= \{\langle a, 1\rangle\}$ since using PK, it costs 2 to delete $a$ and using PK', it costs only 1.

However, as we will see in 2.1.4, if a formula $\varphi$ appears several times in PK then we may replace all the occurrences of the formula $\varphi$ by only one occurrence of $\varphi$ annotated with the sum of the penalties associated to this formula in the previous base. The new knowledge base obtained is equivalent to the initial base.

$\mathcal{B}_c$ will be the set of all the penalty knowledge bases. Note that when the penalties are all infinite, penalty logic comes down to classical logic (no formula can be violated).

Lastly, we will say that PK $\in \mathcal{B}_c$ is consistent if the set of formulas $\varphi_i$ of PK is consistent (without mentioning the penalties $\alpha_i$). Also, in the expressions *sub-theory of* PK, *subset of* PK and PK $\setminus A$ we will refer to the set of formulas obtained from PK by ignoring the penalties.

#### 2.1.1 Cost of an interpretation

Let PK $= \{\langle \varphi_i, \alpha_i \rangle, i = 1 \ldots n\}$ be a penalty knowledge base.

**Definition 1 (Pinkas 91 [17])** *The cost of an interpretation $\omega \in \Omega$ with respect to PK, denoted by $k_{\text{PK}}(\omega)$, is equal to the sum of the penalties of the formulas in PK violated by $\omega$:*

$$k_{\text{PK}}(\omega) = \sum_{\varphi_i \in \text{PK}, \omega \models \neg \varphi_i} \alpha_i$$

*(with the convention $\sum_{\varphi_i \in \emptyset} \alpha_i = 0$ )*

**Definition 2** *A PK-preferred interpretation is an interpretation of minimal cost w.r.t. PK, i.e. an interpretation minimizing $k_{\text{PK}}$.*

As an example, let us consider the following penalty knowledge base $\text{PK}_1$:

$$\begin{array}{lll}
\varphi_1 = a & , & \alpha_1 = +\infty \\
\varphi_2 = b \vee c & , & \alpha_2 = 10 \\
\varphi_3 = \neg b & , & \alpha_3 = 5 \\
\varphi_4 = \neg c & , & \alpha_4 = 7
\end{array}$$

Here are the corresponding interpretations costs:

$$\begin{array}{lll}
k_{\text{PK}_1}(\{\neg a, b, c\}) & = & k_{\text{PK}_1}(\{\neg a, \neg b, c\}) = +\infty \\
k_{\text{PK}_1}(\{\neg a, b, \neg c\}) & = & k_{\text{PK}_1}(\{\neg a, \neg b, \neg c\}) = +\infty \\
k_{\text{PK}_1}(\{a, \neg b, \neg c\}) & = & 10 \\
k_{\text{PK}_1}(\{a, b, \neg c\}) & = & 5 \\
k_{\text{PK}_1}(\{a, \neg b, c\}) & = & 7 \\
k_{\text{PK}_1}(\{a, b, c\}) & = & 5 + 7 = 12
\end{array}$$

If the interpretations are decisions to make (for example if the knowledge base is made of constraints concerning the construction of a timetable), then a minimum cost interpretation corresponds to the cheapest decision, *i.e.*, the most interesting one. The cheapest interpretation is generally not unique. Besides, if the penalties are all equal to 1 then a cheapest interpretation satisfies a maximum consistent subset of PK w.r.t. cardinality.

### 2.1.2 Cost of consistency of a formula

**Definition 3** *The cost of consistency of a formula $\varphi$ with respect to PK, denoted by $K_{PK}(\varphi)$, is the minimum cost with respect to PK of an interpretation satisfying $\varphi$:*

$$K_{PK}(\varphi) = \min_{\omega \models \varphi} k_{PK}(\omega)$$

*(with the convention $\min_\emptyset k_{PK}(\omega) = +\infty$)*

Example:

$$\begin{aligned}
K_{PK_1}(a \wedge b) &= 5 \\
K_{PK_1}(a \rightarrow c) &= 7 \\
K_{PK_1}(\neg a) &= +\infty
\end{aligned}$$

The cost $K_{PK}(\varphi)$ of a formula $\varphi$, is the minimal price to pay in order to make PK consistent with $\varphi$. For example, in order to make $PK_1$ consistent with $a \rightarrow c$, the least expensive way is to remove $\varphi_4$.

**Property 1** $K_{PK}(\bot) = +\infty$ and $K_{PK}(\top) = \min_{\omega \in \Omega}\{k_{PK}(\omega)\}$

All proofs can be found (in French) in Dupin de Saint-Cyr, Lang and Schiex 94 [8] and in Dupin de Saint-Cyr 93 [7].

$K_{PK}(\bot) = +\infty$ is easy to understand, because it is impossible to have PK consistent with $\bot$. Let us note that $K_{PK}(\top)$ is the cost of any PK-preferred interpretation; it is thus the minimum cost to make PK consistent.

**Property 2** $K_{PK}(\top) = +\infty \Leftrightarrow \{\varphi_i \in PK, \alpha_i = +\infty\}$ *is inconsistent.*

This quantity $K_{PK}(\top)$ is important, because it measures the strength of the inconsistency of PK (*i.e.*, how expensive it will be to recover the consistency). If the penalties are all equal to $+\infty$ then $K_{PK}(\top)$ can only take two values: 0 if and only if PK is consistent, and $+\infty$ if and only if PK is inconsistent.

Example: $K_{PK_1}(\top) = 5$; the only minimum cost interpretation is $\{a, b, \neg c\}$. To make $PK_1$ consistent, the least expensive solution is to take off (or to ignore) the formula $\varphi_3$.

**Property 3** $K_{PK}(\top) = 0 \Leftrightarrow PK$ *is consistent.*

Indeed, if $K_{PK}(\top) = 0$ then there is no need to delete any formula in order to make PK consistent, therefore PK is consistent (and conversely).

**Property 4** $\forall \varphi, \psi \in \mathscr{L}, (\varphi \models \psi) \Rightarrow K_{PK}(\varphi) \geq K_{PK}(\psi)$

This property is the monotonicity of K with respect to classical entailment.

**Property 5** $\forall \varphi, \psi \in \mathscr{L}$:

1. $K_{PK}(\varphi \wedge \psi) \geq \max(K_{PK}(\varphi), K_{PK}(\psi))$
2. $K_{PK}(\varphi \vee \psi) = \min(K_{PK}(\varphi), K_{PK}(\psi))$
3. $K_{PK}(\bot) \geq K_{PK}(\varphi) \geq K_{PK}(\top)$

Note that, up to its interval of definition and its ordering convention w.r.t. Proposition 5 $(([0, +\infty], \geq)$ instead of $([0, 1], \leq))$, $K_{PK}$ is actually a possibility measure. Note also that Spohn's ordinal conditional functions $\kappa$ verify property 2 *i.e.* $\kappa(A \cup B) = \min(\kappa(A), \kappa(B))$ [26].

### 2.1.3 Cost of a sub-theory

**Definition 4 (Pinkas 91 [17])** *The cost $C_{PK}(A)$ of a sub-theory $A$ of PK, is the sum of the penalties of the formulas of PK that are not in $A$:*

$$C_{PK}(A) = \sum_{\{\varphi_i, \alpha_i\} \in PK \setminus A} \alpha_i$$

For instance, considering the knowledge base $PK_1$, given $A_1 = \{\varphi_1, \varphi_2, \varphi_3\}$ and $A_2 = \{\varphi_2, \varphi_4\}$, we have $C_{PK_1}(A_1) = \alpha_4 = 7$ and $C_{PK_1}(A_2) = \alpha_1 + \alpha_3 = +\infty$.

**Definition 5** $\forall A, B \subseteq PK$,

$B \geq_{PK} A$ *(B is preferred to A) iff $C_{PK}(B) \leq C_{PK}(A)$.*

$\forall A, B \subseteq PK$, $B >^c_{PK} A$ *if and only if $B \geq_{PK} A$ and not $A \geq_{PK} B$.*

**Definition 6 (Pinkas 91 [17])** $A \subseteq PK$ *is a preferred sub-theory relatively to PK (or $\geq_{PK}$-preferred) if and only if $A$ is consistent and $\nexists B \subseteq PK$, such that $B$ is consistent and $B >^c_{PK} A$.*

Note that there may be several preferred sub-theories (in the previous example, $\{\varphi_1, \varphi_2, \varphi_4\}$ is the only one $\geq_{PK_1}$-preferred sub-theory).

**Property 6** $\forall PK \in \mathscr{B}_c$,

*If $K_{PK}(\top) \neq +\infty$, then any $\geq_{PK}$-preferred sub-theory is a maximal sub-theory of PK w.r.t. inclusion.*

- Let us notice that when $K_{PK}(\top) = +\infty$, every sub-theory of PK has an infinite cost, therefore every sub-theory of PK is $\geq_{PK}$-preferred, but obviously every sub-theory is not necessarily maximal w.r.t. inclusion.

- Besides, if PK is consistent, then $K_{PK}(\top) = 0$, and then the only $\geq_{PK}$-preferred sub-theory of PK is PK itself (its cost is 0).

Example (continued): $A_3 = \{\varphi_1, \varphi_2, \varphi_4\}$ is a $\geq_{PK_1}$-preferred sub-theory and it is maximal w.r.t. inclusion.

But, although $\{\varphi_2, \varphi_3, \varphi_4\}$ is a maximal sub-theory of $PK_1$ (w.r.t. inclusion), it is not $\geq_{PK_1}$-preferred (because its cost is infinite).

If we add the formula $\langle \varphi_5 = \neg a, \alpha_5 = +\infty \rangle$ to $PK_1$ then the subset of infinite cost formulas is inconsistent, therefore every sub-theory has an infinite cost, and every sub-theory is a preferred sub-theory.

**Property 7** *The cost $k_{PK}(\omega)$ of an interpretation $\omega \in \Omega$ with respect to PK is equal to the cost of the sub-theory of PK composed of all the formulas satisfying $\omega$:*

$$\forall \omega \in \Omega, k_{PK}(\omega) = C_{PK}(PK \cap [\omega])$$

**Corollary 7.1** $\forall \omega \in \Omega$,

$\omega$ *has a minimal cost w.r.t.* PK
$\Leftrightarrow$
PK $\cap [\omega]$ *is a $\geq_{PK}$-preferred sub-theory with respect to all the sub-theories of* PK

**Corollary 7.2** *A is a maximal sub-theory of* PK $\Rightarrow$ $\forall \omega \models A, k_{PK}(\omega) = C_{PK}(A)$.

**Corollary 7.3** $K_{PK}(\varphi)$ *is equal to the minimum cost of a $\varphi$-consistent sub-theory of* PK*:*

$$K_{PK}(\varphi) = \min_{A \subseteq PK, A\ \varphi-consistent} C_{PK}(A)$$

Therefore, the cost of a formula $\varphi$ with respect to the base PK is the cost of a $\varphi$-consistent $\geq_{PK}$-preferred sub-theory of PK.

**Corollary 7.4** $\forall A \subseteq PK$,

*A is a $\geq_{PK}$-preferred sub-theory*
$\Leftrightarrow K_{PK}(\top) = C_{PK}(A)$.

(cf. corollary 7.3, with $\varphi = \top$).

**Definition 7** $Add(PK, \varphi) = PK \cup \{\langle \varphi, +\infty \rangle\}$

**Property 8**

$$K_{PK}(\varphi) = K_{Add(PK, \varphi)}(\top)$$

Therefore, the cost to make the knowledge base consistent with a given formula, can be computed by adding this formula with an infinite penalty and then evaluating the cost of the new knowledge base consistency.

### 2.1.4 Equivalence between penalty knowledge bases

Two penalty knowledge bases are *semantically equivalent* if they induce the same cost function on $\Omega$, i.e.:

**Definition 8** $\forall PK, PK' \in \mathscr{B}_c$,

PK $\approx^c$ PK' *(PK is semantically equivalent to PK')*
$\Leftrightarrow k_{PK} = k_{PK'}$.

Besides, we define a pre-ordering relation $\ll^c$ on $\mathscr{B}_c$ as follows:

**Definition 9** $\forall PK, PK' \in \mathscr{B}_c$,

PK $\ll^c$ PK' *(PK is less expensive than PK')*
$\Leftrightarrow k_{PK} \leq k_{PK'}$

As an example, let us consider $PK_2$, $PK_3$ and $PK_4$ the penalty knowledge bases defined as follows:

| $PK_2$: | $PK_3$: | $PK_4$: |
|---|---|---|
| a  5 | a  8 | $a \wedge b$  18 |
| a  3 | b  10 | |
| b  10 | | |

The cost functions induced by those bases are the following:

| $\omega$ | $k_{PK_2}(\omega)$ | $k_{PK_3}(\omega)$ | $k_{PK_4}(\omega)$ |
|---|---|---|---|
| $\{a, b\}$ | 0 | 0 | 0 |
| $\{a, \neg b\}$ | 10 | 10 | 18 |
| $\{\neg a, b\}$ | 8 | 8 | 18 |
| $\{\neg a, \neg b\}$ | 18 | 18 | 18 |

So we have $PK_2 \approx^c PK_3$ and $PK_3 \ll^c PK_4$ (but $PK_3$ is not equivalent to $PK_4$).

N.B.: the previous example shows that it is impossible to transform equivalently a penalty knowledge base containing several non-equivalent formulas in a penalty knowledge base containing the conjunction of those formulas.

*But, if a knowledge base contains several times the same formula (or an equivalent one), it is possible to transform it equivalently in a knowledge base containing this formula only one time with a penalty equal to the sum of the penalties of this formula in the previous base.*

**Property 9** $\forall PK, PK' \in \mathscr{B}_c$,

$$PK \approx^c PK' \Rightarrow \wedge \{\varphi_i | \varphi_i \in PK\} \models \wedge \{\varphi_j | \varphi_j \in PK'\}$$

The converse is obviously false.

### 2.2 Inconsistency handling with penalty logic

Using penalties to handle inconsistency is a *syntax-based approach*, in the sense of [16], which means that the way a knowledge base behaves is dependent on the syntax of the input (this is justified by the fact that each formula is considered as an independent piece of information); for instance, $\{p, q, \neg p \vee \neg q\}$ will not behave as $\{p \wedge q, \neg p \vee \neg q\}$, since in the first case we can remove independently the formulas $p$ and $q$ ($\{p, q\}$, $\{p, \neg p \vee \neg q\}$ and $\{q, \neg p \vee \neg q\}$ are the maximal sub-theories), but in the second case we must remove or keep the *whole* formula $p \wedge q$ ($\{p \wedge q\}$ and $\{\neg p \vee \neg q\}$ are the maximal sub-theories).

In order to deal with inconsistency, the basic idea developed with syntax-based approaches is to define a nonmonotonic inference relation as follows: $\psi$ can be deduced nonmonotonicaly from a knowledge base iff all the maximal sub-theories of this base entails (classically) $\psi$.

#### 2.2.1 Nonmonotonic inference relation induced by a penalty knowledge base

Given $\mathrm{PK} \in \mathscr{B}_c$.

**Definition 10** $\forall \varphi, \psi \in \mathscr{L}$,
$$\varphi \vdash^c_{\mathrm{PK}} \psi \Leftrightarrow$$
$\forall A \subseteq \mathrm{PK}$, if $A$ is a $\geq_{\mathrm{PK}}$-preferred $\varphi$-consistent sub-theory among all the $\varphi$-consistent sub-theories of $\mathrm{PK}$, then $A \cup \{\varphi\} \models \psi$.

In particular, if $\varphi = \top$, the definition becomes: $\vdash^c_{\mathrm{PK}} \psi \Leftrightarrow$ if $A$ is a $\geq_{\mathrm{PK}}$-preferred sub-theory among all the sub-theories of PK, then $A \models \psi$.

N.B.: $\forall \psi, \bot \vdash^c_{\mathrm{PK}} \psi$.

**Property 10** $\forall \varphi, \psi \in \mathscr{L}$,
$$\varphi \vdash^c_{\mathrm{PK}} \psi \Leftrightarrow$$
$\forall \omega \in \Omega$, if $\omega \models \varphi$ and $\omega$ is a $\geq_{\mathrm{PK}}$-preferred interpretation satisfying $\varphi$, then $\omega \models \psi$.

This property shows that the nonmonotonic inference relation $\vdash^c_{\mathrm{PK}}$ belongs to the set of relations based on preferential models in the sense of [15]. As $\geq_{\mathrm{PK}}$ is a complete pre-ordering, we immediately get the following result:

**Property 11** $\vdash^c_{\mathrm{PK}}$ is a comparative inference relation[1].

**Property 12** Given $\mathrm{PK} \in \mathscr{B}_c$ and $\varphi, \psi \in \mathscr{L}$, with $\varphi \neq \bot$,
$$\varphi \vdash^c_{\mathrm{PK}} \psi \Leftrightarrow \varphi \models \psi \text{ or } \vdash^c_{Add(\mathrm{PK},\varphi)} \psi$$

For instance, let us consider the following penalty knowledge base PK:
$$\{\langle a \vee b, +\infty \rangle,$$
$$\langle \neg a, 5 \rangle,$$
$$\langle \neg a \vee \neg b, 4 \rangle,$$
$$\langle b \to \neg c, 2 \rangle,$$
$$\langle a \to c, 1 \rangle\}$$

It can be checked that:
$$\vdash^c_{\mathrm{PK}} \neg c$$
$$a \vdash^c_{\mathrm{PK}} c$$
$$(a \wedge b) \vdash^c_{\mathrm{PK}} \neg c$$

---
[1] A comparative inference relation [13] is a rational relation [15] that also satisfies supraclassicality: if $\varphi \models \psi$ then $\varphi \vdash^c_{\mathrm{PK}} \psi$.

### 2.3 An application of penalty logic: maximum clique in a graph

In this section, we will see that penalty logic is not only a tool for inconsistency handling but also a good way to represent, in a logical language, discrete optimization problems (for instance issued from operation research), in which minimum cost interpretations correspond to optimum solutions.

We consider an undirected graph $G$, i.e., a set of vertices $U$ and a set of edges $V$ connecting those vertices. A clique of $G$ is a subset of $V$ which define a complete sub-graph (i.e., every vertex is connected with every other vertex). Finding a maximum cardinality clique is a classical NP-hard problem in operational research. In penalty logic we can represent it like this:

- to each vertex $s \in U$, we can associate a propositional variable $s$ which truth assignment means that this vertex *belongs* to the clique we are looking for.

- we are searching for a set of vertices which is maximum for cardinality, so we have to exclude the minimum of vertices: to each vertex we associate the penalty formula $\langle s, 1 \rangle$.

- the resulting set must be a clique so for each pair $(x, y)$ of vertices that are not connected in the graph $G$ (i.e., $(x, y) \notin V$), at least either $x$ or $y$ does not belong to the clique. In consequence, we can associate to each pair $(x, y) \notin V$ the penalty formula $\langle \neg x \vee \neg y, +\infty \rangle$.

Let $\mathrm{PK}(G) = \{\langle s, 1 \rangle, s \in U\} \cup \{\langle \neg x \vee \neg y, +\infty \rangle, (x, y) \notin V\}$.

**Property 13** (see [8]) *Every minimum cost interpretation with respect to $\mathrm{PK}(G)$ corresponds to a maximum clique of $G$ and conversely.*

Example:

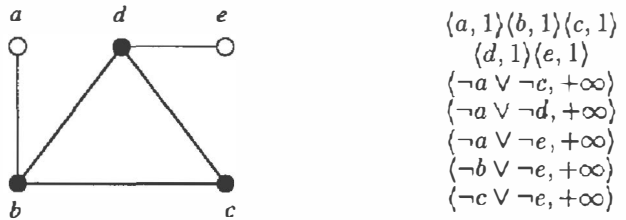

$\langle a, 1 \rangle \langle b, 1 \rangle \langle c, 1 \rangle$
$\langle d, 1 \rangle \langle e, 1 \rangle$
$\langle \neg a \vee \neg c, +\infty \rangle$
$\langle \neg a \vee \neg d, +\infty \rangle$
$\langle \neg a \vee \neg e, +\infty \rangle$
$\langle \neg b \vee \neg e, +\infty \rangle$
$\langle \neg c \vee \neg e, +\infty \rangle$

The minimum cost interpretation is $\{\neg a, b, c, d, \neg e\}$.

This example shows the ability of penalty logic to encode discrete optimization problems. One could argue that, in operation research, algorithms for solving classical problems (as maximum clique, minimum vertex cover...) do already exist. Those algorithms are probably more efficient than the one consisting in finding the best interpretation in penalty logic (developed in [7]). However, the logical representation of this kind of problems presents at least two advantages: the great power of expression of logic allows us to specify many complicated problems which could not easily be specified within the operational research language; and the best solution search method is independent of the given problem.

## 3  Relating penalties to Dempster-Shafer theory

In this section we are going to show:

- first, that the cost of an interpretation $k_{PK} : \Omega \to [0, +\infty]$ induced by a penalty knowledge base PK consisting of $n$ weighted formulas corresponds actually to the contour function $pl : \Omega \to [0, 1]$ induced by Dempster's combination of $n$ simple support functions (one for each formula $\varphi_i$);

- then, that moreover, the function $K_{PK} : \mathcal{L} \to [0, +\infty]$ corresponds to a plausibility measure in an *infinitesimal* version of Depmster-Shafer theory.

### 3.1  Interpretation costs and contour functions

Let PK $= \{\langle\varphi_i, \alpha_i\rangle, i = 1\ldots n\}$ be a penalty knowledge base. Let us define, for each $i$, the body of evidence $m_i$:

$$m_i(\varphi_i) = 1 - e^{-\alpha_i}$$
$$m_i(\top) = e^{-\alpha_i}$$

By convention we take $e^{-\infty} = 0$. Since $\alpha_i \in [0, +\infty]$, it can be seen that $m_i(\varphi_i) \in (0, 1]$ and $m_i(\top) \in [0, 1)$. Moreover, note that $\lim_{\alpha_i \to +\infty} m_i(\varphi_i) = 1$. $m_i$ is called a simple support function [24]. Let $m = m_1 \oplus \cdots \oplus m_n$ be the result of Dempster's combination of the $m_i$'s [9] without re-normalization. The contour function $pl : \Omega \to [0, 1]$ associated to $m$ is the restriction of the plausibility function to singletons, *i.e.*,

$$pl(\omega) = Pl(\{\omega\}) = \sum_{\varphi, \omega \models \varphi} m(\varphi)$$

Now, it is well-known [24] that

$$Pl(\{\omega\}) = \prod_{i=1}^{n} Pl_i(\{\omega\})$$

where $Pl_i$ is the plausibility function induced by $m_i$. Moreover, $Pl_i(\omega) = 1$ if $\omega \models \varphi_i$ and $Pl_i(\omega) = e^{-\alpha_i}$ if $\omega \models \neg\varphi_i$. Thus,

$$\begin{aligned}
pl(\omega) &= (\prod_{i,\omega\models\varphi_i} 1) \cdot (\prod_{i,\omega\models\neg\varphi_i} e^{-\alpha_i}) \\
&= \prod_{i,\omega\models\neg\varphi_i} e^{-\alpha_i} \\
&= e^{-\sum_{i,\omega\models\neg\varphi_i} \alpha_i} \\
&= e^{-k_{PK}(\omega)}
\end{aligned}$$

Therefore, $k_{PK}(\omega) = -\ln(pl(\omega))$: up to a logarithmic transformation, $k_{PK}$ is a contour function, or more precisely, the process consisting in computing $k_{PK}$ corresponds to applying Dempster's combination without re-normalization on simple support functions. This equivalence does not extend to an equivalence between $K_{PK}$ and a plausibility function (see subsection 3.2), but this result is already significant, since in most practical applications of penalty logic, only the contour function $k_{PK}$ is useful: this is the case when penalties are used to induce a preference relation on $\Omega$, and then possibly to select one of the (or all) cheapest interpretation(s). Namely, this is enough for inducing the inference relation $\vdash^c_{PK}$, for solving discrete optimization problems, and also for applying penalties to constraint satisfaction problems or abduction. So, handling penalties in such a purpose is nothing but performing Dempster's combination on simple support functions. Reciprocally, combining simple support functions in order to rank interpretations can be done alternatively with penalty logic.

This also brings to light a relation between penalties and [25] where each formula $\varphi_i$ of the knowledge base is considered to be given by a *distinct source*, this source having the probability $p_i$ to be faulty (*i.e.*, the information it provides us with is not pertinent), and all sources being independent (which gives the simple support function $m_i(\varphi_i) = (1 - p_i)$ and $m_i(\top) = p_i$). So if the task is only to find the most plausible interpretation (as in [11] which is the Constraint Satisfaction counterpart of [25]), it can thus be done equivalently with penalties.

### 3.2  Formula costs as infinitesimal plausibilities

Let us consider an infinitesimal version of Dempster-Shafer theory, where the masses involved are all infinitely close to 0 or to 1. Let $\varepsilon$ be an infinitely small quantity[2,3]. Again, let PK $= \{\langle\varphi_i, \alpha_i\rangle, i = 1\ldots n\}$. Let us define, for each $i$, the infinitesimal body of evidence $m_{\varepsilon,i}$:

$$m_{\varepsilon,i}(\varphi_i) = 1 - \varepsilon^{\alpha_i}$$
$$m_{\varepsilon,i}(\top) = \varepsilon^{\alpha_i}$$

Let $m_\varepsilon = m_{\varepsilon,1} \oplus \cdots \oplus m_{\varepsilon,n}$ be the result of Dempster's combination of the $m_i$'s [9] without re-normalization. Let us show now that $K_{PK}$ has the same order of magnitude (w.r.t. $\varepsilon$) as $\ln(Pl_\varepsilon)$, where $\ln(Pl_\varepsilon)$ is the plausibility function induced by $m_\varepsilon$.

Let us note that the set of focal elements of $m_\varepsilon$ is exactly $\{\wedge_{i\in I}\varphi_i, I \subseteq \{1\ldots n\}\}$.

---

[2]More formally, this consists in considering a family of $\varepsilon$'s tending towards 0; indeed what we are interested in is only the limit of the considered $f(\varepsilon)$ when $\varepsilon$ tends to 0.

[3]We recall that $f_1(\varepsilon) \approx f_2(\varepsilon)$ iff $\lim_{\varepsilon \to 0} \frac{f_1(\varepsilon)}{f_2(\varepsilon)} = 1$.

Now, let us define

$$R(PK, \psi) = \{I \subseteq \{1 \ldots n\}, \bigwedge_{i \in I}(\varphi_i) \wedge \psi \; consistent\}$$

Now,

$$\begin{aligned} Pl(\psi) &= \sum_{I \in R(PK,\psi)} \prod_{i \in I} m_i(\varphi_i) . \prod_{i \notin I} m_i(\top) \\ &= \sum_{I \in R(PK,\psi)} \prod_{i \in I}(1 - \varepsilon^{\alpha_i}) . \prod_{i \notin I} \varepsilon^{\alpha_i} \end{aligned}$$

As $\varepsilon$ is infinitely small and $I$ is always finite, $\prod_{i \in I}(1 - \varepsilon^{\alpha_i}) \approx 1$, therefore:

$$\begin{aligned} Pl(\psi) &\approx \sum_{I \in R(PK,\psi)} \prod_{i \notin I} \varepsilon^{\alpha_i} \\ &\approx \sum_{I \in R(PK,\psi)} \varepsilon^{\sum_{i \notin I} \alpha_i} \end{aligned}$$

Let us now define $R_{minpen}(PK, \psi)$ as

$$\{J \in R(PK, \psi), \sum_{i \notin J} \alpha_i \; is \; minimum\}$$

and let $r(PK, \psi) = |R_{minpen}(PK, \psi)|$.

Since $\varepsilon$ is infinitely small, we have

$$\begin{aligned} Pl(\psi) &\approx \sum_{I \in R_{minpen}(PK,\psi)} \varepsilon^{\sum_{i \notin I} \alpha_i} \\ &\approx r(PK,\psi) . max_{I \in R(PK,\psi)} \varepsilon^{\sum_{i \notin I} \alpha_i} \\ &\approx r(PK,\psi) . \varepsilon^{min_{I \in R(PK,\psi)} \sum_{i \notin I} \alpha_i} \end{aligned}$$

Now,

$$\begin{aligned} min_{I \in R(PK,\psi)} \sum_{i \notin I} \alpha_i &= \min_{B \subseteq PK, B \wedge \psi \; consistent} \sum_{\varphi_i \notin B} \alpha_i \\ &= \min_{B \subseteq PK, B \wedge \psi \; consistent} C_{PK}(B) \\ &= K_{PK}(\psi) \end{aligned}$$

Therefore,

$$Pl(\psi) \approx r(PK, \psi) . \varepsilon^{K_{PK}(\psi)}$$

Note that $r(PK, \psi)$ does not depend on $\varepsilon$, and moreover that $r(PK, \psi) > 0$. So, up to a logarithmic transformation and a multiplicative constant (in other terms, *if we consider only the orders of magnitude w.r.t. $\varepsilon$*), $K_{PK}$ is equivalent to an infinitesimal plausibility function.

## 4  Conclusion

Used to handle inconsistency and perform non-monotonic inferences, penalty logic has shown to have interesting properties. Using penalties for selecting preferred sub-theories of an inconsistent knowledge base not only allows to distinguish between the degree of importance of various formulas, as usual priority-based approaches do, but also to express possible *compensations* between formulas. The non-monotonic inference relation defined satisfies the usual postulates [13] and is (logarithmically) related to an infinitesimal version of Dempster-Shafer theory.

Furthermore, the complexity of the penalty non-monotonic deduction problem has been considered in [5] and is ranked as one of the most simple non-monotonic inference problem (in $\Delta_2^p$).

Penalty logic may also been considered as a logical language for expressing discrete optimization problems. The search for a preferred interpretation has been implemented using an $A^*$-like variant of Davis and Putnam procedure [6] and has been tested on small examples. Randomized search algorithms such as GSAT [23, 22] could also be considered, but they do not guarantee that an optimum is actually reached.

As shown in [5], solving the problem of searching a preferred interpretation allows to simply solve the non-monotonic inference problem, without any restriction on the language of the formulas expressed[4]. Anyway, even the limited $\Delta_2^p$ complexity can be considered as excessive when faced to practical applications. A reasonable approach would then consist in defining a gradual inference relation and in trying only to solve an approximation of the resulting gradual inference problem.

Among the other possible extensions of penalty logic, one could consider associating many unrelated penalties to a single formula. Partially ordered penalty vectors would then replace penalties. Another possible extension consists in taking into account not only penalties caused by violations but also profits associated to satisfactions (which could be expressed using negative penalties).

## Acknowledgements

We would like to express our thanks to Didier Dubois and Henri Prade for helpful suggestions concerning the link between penalties and Dempster-Shafer theory, and Michel Cayrol for having found an error in a preliminary version. This work has been partially supported by the ESPRIT BRA project DRUMS-2.

---

[4]Using an ATMS for computing candidates and preferred sub-theories could also be considered, but the resulting complexity is more important in the general case [19].